%% file: root.tex
\documentclass[letterpaper, 10 pt, conference]{ieeeconf}

\usepackage{bbm}
\usepackage{cite}
\usepackage{url}
\usepackage{comment}
\usepackage{algorithmic}
\usepackage{graphicx}
\usepackage{textcomp}
\usepackage{xcolor}
\usepackage{amsmath,amssymb,amsfonts}

\usepackage{multirow}
\usepackage{booktabs}

\usepackage{tikz}
\usepackage{makecell}
\usepackage{setspace}
\usepackage{stfloats}
\usepackage{lipsum}
\usepackage{bbding}
\usepackage[colorlinks,bookmarksopen,bookmarksnumbered,citecolor=green, linkcolor=red, urlcolor=black]{hyperref}

\include{shortcuts}

\include{acronyms}

\IEEEoverridecommandlockouts                              
\title{\LARGE \bf
  Advancing Off-Road Autonomous Driving: The Large-Scale ORAD-3D Dataset and Comprehensive Benchmarks
}

\author{Chen Min$^{1}$, Jilin Mei$^{1}$, Heng Zhai$^{2}$, Shuai Wang$^{1}$, Tong Sun$^{1}$, Fanjie Kong$^{3}$, Haoyang Li$^{4}$, Fangyuan Mao$^{1}$, \\Fuyang Liu$^{1}$, Shuo Wang$^{1}$, Yiming Nie$^{5}$, Qi Zhu$^{5}$, Liang Xiao$^{5}$, Dawei Zhao$^{5}$, Yu Hu$^{1}$
\thanks{$^{1}$Research Center for Intelligent Computing Systems, SKLP, Institute of Computing Technology, Chinese Academy of Sciences, Beijing, China, 100190.}%
\thanks{$^{2}$Tongji University, Shanghai, China, 200092.}%
\thanks{$^{3}$Xi'an Jiaotong University, Shaanxi, China, 710049.}%
\thanks{$^{4}$Nanchang University, Jiangxi, China, 330047.}%
\thanks{$^{5}$Defense Innovation Institute, Beijing, China, 100073.}%
\thanks{$^{*}$ Corresponding author Dawei Zhao and Yu Hu. Email: {\tt\small adamzdw@163.com and huyu@ict.ac.cn}}%
}

\begin{document}

\maketitle
\thispagestyle{empty}
\pagestyle{empty}

\begin{abstract}
A major bottleneck in off-road autonomous driving research lies in the scarcity of large-scale, high-quality datasets and benchmarks. To bridge this gap, we present ORAD-3D, which, to the best of our knowledge, is the largest dataset specifically curated for off-road autonomous driving. ORAD-3D covers a wide spectrum of terrains—including woodlands, farmlands, grasslands, riversides, gravel roads, cement roads, and rural areas—while capturing diverse environmental variations across weather conditions (sunny, rainy, foggy, and snowy) and illumination levels (bright daylight, daytime, twilight, and nighttime). Building upon this dataset, we establish a comprehensive suite of benchmark evaluations spanning five fundamental tasks: 2D free-space detection, 3D occupancy prediction, rough GPS-guided path planning, vision–language model–driven autonomous driving, and world model for off-road environments. Together, the dataset and benchmarks provide a unified and robust resource for advancing perception and planning in challenging off-road scenarios. The dataset and code will be made publicly available at \url{https://github.com/chaytonmin/ORAD-3D}.

\end{abstract}

\section{Introduction}

\begin{figure}
	\centering
	\begin{minipage}[b]{.8\linewidth}
		\centerline{\includegraphics[width=8cm]{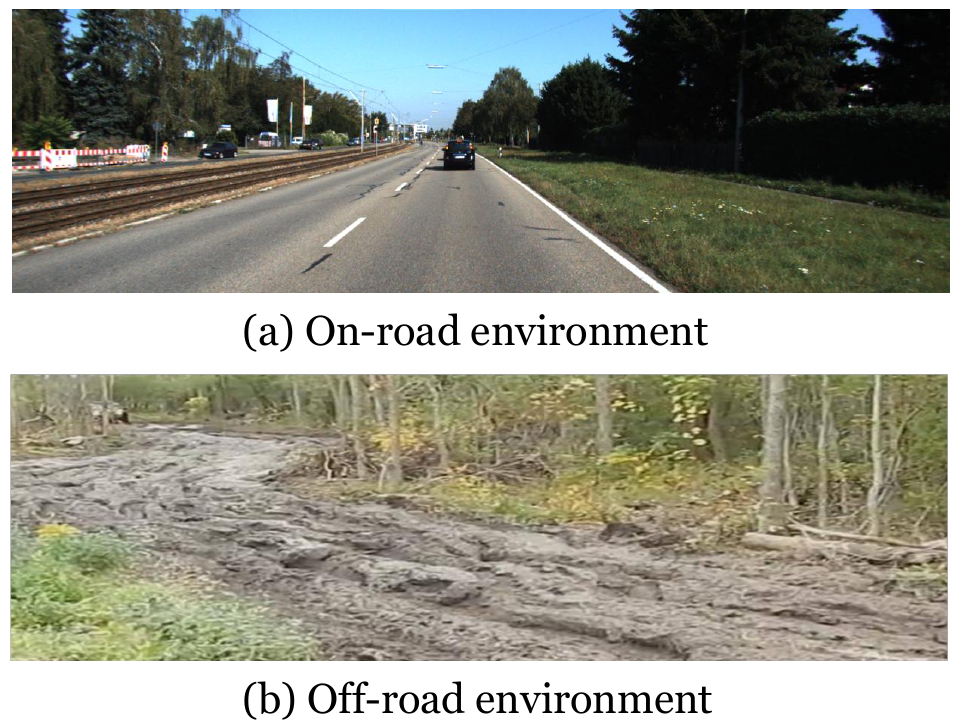}}
	\end{minipage}
	\caption{On-road environments are characterized by well-structured infrastructure—such as lane markings, traffic signs, and clearly delineated roadways—that provide consistent and reliable navigational cues. In stark contrast, off-road environments lack paved surfaces and structured guidance, instead featuring irregular, heterogeneous, and often ambiguous terrain. This absence of formal infrastructure greatly amplifies the complexity of perception and navigation tasks.}
	\label{difference}
\end{figure}

Off-road autonomous driving has attracted increasing attention in recent years due to its potential in enabling intelligent transportation and robotic systems to operate in unstructured and complex environments~\cite{min2024autonomous}. As illustrated in Fig.~\ref{difference}, unlike urban or highway settings—where well-marked lanes, standardized traffic rules, and high-quality maps provide strong priors—off-road scenarios are characterized by irregular terrains, sparse or absent road markings, unpredictable obstacles, and rapidly changing environmental conditions~\cite{shu2025overview}. These factors pose significant challenges for perception, planning, and control systems, necessitating robust algorithms capable of handling extreme variability in both appearance and geometry~\cite{chen2025scene}.

Despite the growing interest in this domain, progress in off-road autonomous driving research has been hindered by the scarcity of large-scale, high-quality datasets. Existing datasets for autonomous driving predominantly focus on structured, on-road environments, limiting the generalization and robustness of models when deployed off-road. Moreover, the few publicly available off-road datasets are often constrained in size, diversity, or sensing modalities, leaving a gap in supporting comprehensive research on perception and planning under diverse off-road conditions.

To address this gap, we present ORAD-3D—to the best of our knowledge, the largest and most diverse dataset dedicated to off-road autonomous driving. The dataset was collected across a broad spectrum of terrains, including woodlands, farmlands, grasslands, riversides, gravel roads, cement roads, and rural areas. It encompasses a rich set of environmental conditions, covering multiple weather scenarios (sunny, rainy, foggy, and snowy) and illumination states (bright daylight, daytime, twilight, and nighttime). The data are captured using multi-sensor configurations, enabling both 2D and 3D perception tasks.

In addition to releasing ORAD-3D, we establish a comprehensive set of benchmark evaluations to enable fair and reproducible comparisons in off-road autonomous driving research. Specifically, we evaluate five core tasks: (1) 2D free-space detection, (2) 3D occupancy prediction, (3) rough GPS-guided path planning, (4) VLM-based autonomous driving, and (5) world model for off-road autonomous driving.
Determining traversable areas is essential for safe off-road navigation; thus, we first construct a benchmark for 2D free-space detection based on large-scale off-road imagery. Recognizing the critical role of 3D terrain geometry, we further provide 3D occupancy annotations for predicting terrain structure in 3D space. Given that GPS signals in off-road environments are often noisy or inaccurate—making precise localization challenging—we design a benchmark for rough GPS-guided path planning to evaluate navigation under such uncertainty. Inspired by the rapid progress and strong generalization capability of Vision–Language Models (VLMs), we introduce a benchmark for VLM-based scene understanding and path planning in diverse off-road scenarios. Finally, we construct an off-road world model capable of controllably generating diverse future off-road scenarios.
Together, these benchmarks span a representative spectrum of perception and decision-making challenges encountered in real-world off-road environments.

By providing both a large-scale, diverse dataset and well-defined benchmarks, ORAD-3D aims to serve as a foundational resource for the off-road autonomous driving community. We anticipate that it will accelerate the development of algorithms that are robust to terrain variability, adverse weather, and low-visibility conditions—ultimately pushing the frontier of safe and reliable off-road autonomous driving.

The highlights of our work are as follows:
\begin{itemize}
	\item \textbf{ORAD-3D dataset} – We release the largest and most diverse off-road autonomous driving dataset to date, covering multiple terrains, weather conditions, and illumination levels with synchronized multi-sensor data.
	
	\item \textbf{Environmental diversity} – The dataset includes challenging and underrepresented scenarios such as woodlands, farmlands, grasslands, riversides, and rural roads, with extensive weather and lighting variations.
	
	\item \textbf{Benchmark suite} – We establish standardized benchmarks for 2D free-space detection, 3D occupancy prediction, rough GPS-guided path planning, VLM-based autonomous driving, and off-road world model.
	
	\item \textbf{Research resource} – ORAD-3D serves as a comprehensive platform for developing and evaluating robust perception and planning algorithms in unstructured off-road environments.
\end{itemize}

\section{Related Work}

\subsection{Off-Road Autonomous Driving}
Off-road perception research primarily targets road segmentation and traversability estimation in unstructured, complex terrains. Recent works enhance passable area detection using advanced CNNs and Transformer architectures~\cite{sun2023passable,chung2024pixel}, while others integrate RGB imagery with point clouds to achieve more robust terrain mapping and navigation~\cite{guan2021tns,guan2023vinet}.
Multi-modal fusion techniques combining RGB, depth, and LiDAR data have been employed to improve free-space detection and obstacle classification~\cite{bae2023self,feng2023adaptive,m2f2,kim2024ufo}. Noise-robust networks and Transformer-based models further refine segmentation accuracy in challenging conditions~\cite{lv2024noise,yan2024fsn}. Contrastive and self-supervised learning approaches reduce dependence on dense labels, enabling fine-grained terrain understanding from limited annotations~\cite{gao2021fine,seo2023learning,jung2023v}.
Off-road semantic segmentation methods such as OFFSEG~\cite{offseg} and OffRoadTranSeg~\cite{singh2021offroadtranseg} tackle class imbalance and domain adaptation issues via semi-supervised strategies. Fusion-based frameworks leverage uncertainty modeling and attention mechanisms to enhance mapping and perception in unstructured environments~\cite{lian2023research,feng2024multi,kim2024uncertainty}.
Xu et al.~\cite{xu2022trajectory} propose an end-to-end Transformer-based framework for map-less autonomous driving, which takes raw LiDAR data and a noisy topometric map as inputs and generates precise local trajectories for navigation.
While these approaches demonstrate promising results, most are trained and evaluated on small-scale datasets. To address this limitation, we introduce a large-scale off-road dataset with comprehensive benchmark results.
\subsection{Datasets for Off-Road Autonomous Driving}

\begin{table}[t]
	\caption{Overview of off-road autonomous driving dataset.}
	\begin{center}
		\resizebox{0.5\textwidth}{!}{
			\begin{tabular}{c|c|c|c|c}
				\hline
				Datasets&Sensors & Frames & Tasks &Extreme Weathers  \\ 
				\hline
				DeepScene~\cite{valada2016deep}& C&0.3K&Seg&	no\\
				RUGD~\cite{rugd} & C&7.5K&Seg&	no\\
				Rellis-3D~\cite{rellis-3d} & C+L&13K&3DDet, Seg&	no\\
				GOOSE~\cite{mortimer2023goose}& C+L&	10K&Seg&	no\\
				ORFD~\cite{orfd} & C+L&	12K&Seg&\makecell{Rain, snow, fog,\\ darkness, grassland, forest, \\ rural area, etc.}\\
				\midrule
				ORAD-3D &C+L&57K &\makecell{Seg, 3D Occ, Path Planning, \\ VLM, World Model} &\makecell{Rain, snow, fog,\\ darkness, grassland, forest, \\ rural area, etc.}\\ 
				\hline
			\end{tabular}
			\label{over}
		}
	\end{center}
\end{table}
Existing off-road autonomous driving datasets predominantly target perception-oriented tasks, such as traversability estimation and semantic segmentation. Identifying traversable areas is particularly critical for autonomous vehicles. TrailNet\cite{hoveidar2018autonomous} is among the first datasets to examine road surface types using publicly available camera data. ORFD\cite{orfd} provides high-resolution imagery and annotations for detecting navigable regions across diverse off-road conditions. Verti-Wheelers\cite{datar2023toward} addresses wheeled mobility on steep terrain, while M3-GMN\cite{m3gmn} advances grid map–based navigation.
For semantic segmentation in off-road environments, several datasets have been introduced. YCOR\cite{maturana2018real} offers multi-season imagery, whereas BotanicGarden\cite{liu2024botanicgarden} contains annotated images for robot navigation in botanical gardens. RUGD\cite{rugd} provides multimodal sensor data to support perception in outdoor scenes. More recently, RELLIS-3D\cite{rellis-3d}, GOOSE\cite{mortimer2023goose}, and WildScenes\cite{wildscenes} have released multimodal datasets for robust perception in complex natural environments. UnScenes3D~\cite{chen2025scene} focuses on 3D semantic occupancy as a central representation for off-road understanding.
While these datasets have significantly contributed to the field, most remain limited in scale and primarily emphasize perception tasks. In contrast, this work introduces a large-scale off-road dataset that not only supports perception tasks but also includes planning tasks and vision–language model–driven autonomy, thereby providing a more comprehensive resource for advancing off-road autonomous driving research.

\begin{figure*}[t]
	\centering
	\centerline{\includegraphics[width=6.8in]{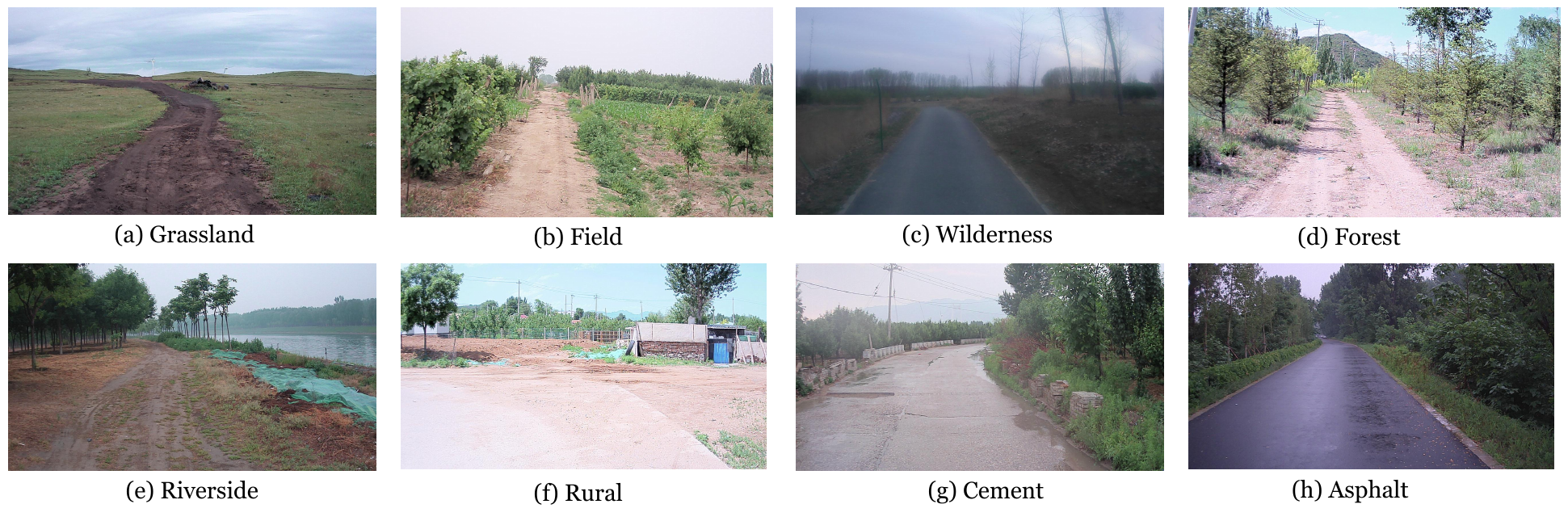}}
	\caption{ORAD-3D dataset contains a variety of off-road scenes.}
	\label{type}
\end{figure*}

\begin{figure}[t]
	\centering
	\centerline{\includegraphics[width=3.4in]{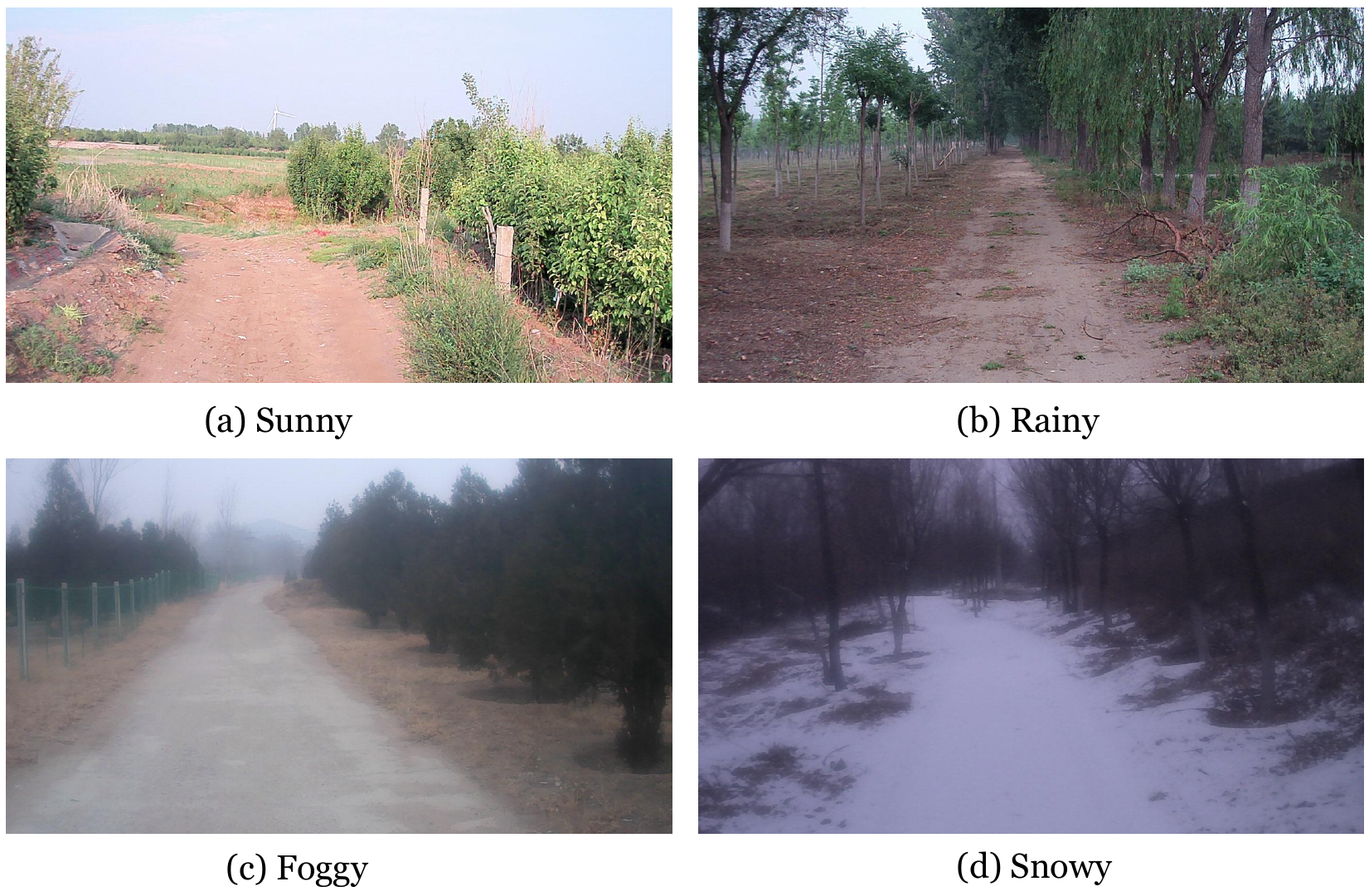}}
	\caption{Different weather conditions are considered in ORAD-3D.}
	\label{weather}
\end{figure}

\begin{figure}[t]
	\centering
	\centerline{\includegraphics[width=3.4in,height=2.2in]{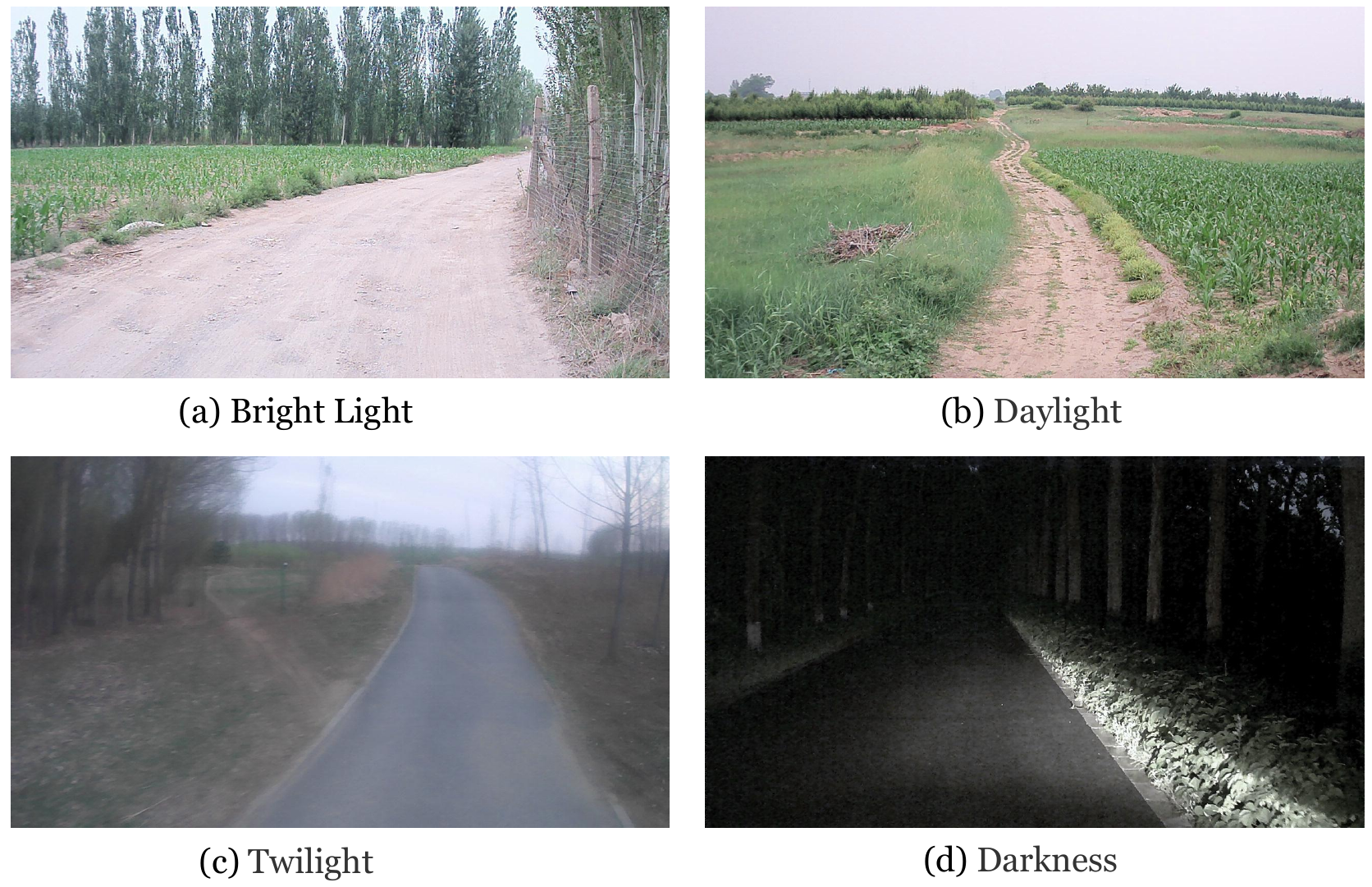}}
	\caption{Different light conditions are considered in ORAD-3D.}
	\label{day}
\end{figure}

\begin{figure}[t]
	\centering
	\centerline{\includegraphics[width=3.4in,height=2.2in]{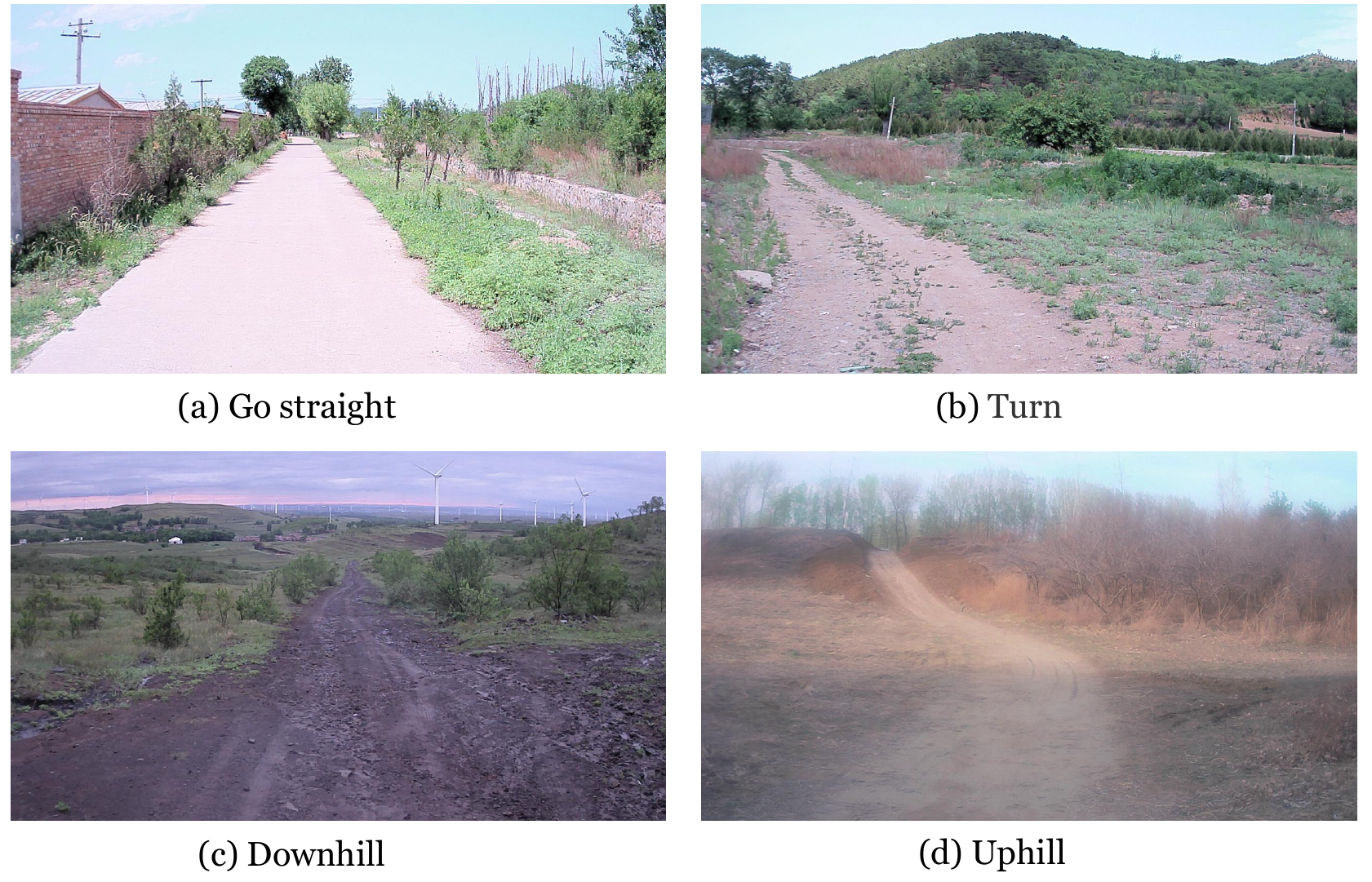}}
	\caption{Different road types are considered in ORAD-3D.}
	\label{turn}
\end{figure}

\section{ORAD-3D Dataset}

We present ORAD-3D, a comprehensive off-road autonomous driving dataset curated from multi-season recordings spanning winter to summer. It covers diverse terrains, weather, lighting, and road types, providing rich data essential for robust perception and accurate future forecasting. ORAD-3D includes data such as images, LiDAR, pose, and depth. The provided annotations include 2D free-space segmentation, 3D occupancy, driving trajectories, and scene text descriptions.

\begin{table*}[t]
	\caption{Distribution of off-road scenes in the ORAD-3D dataset. The unit is frames.}
	\begin{center}
		\resizebox{1\textwidth}{!}
		{
			\begin{tabular}{c|cccccccccc|c|c}
				\hline
				Split&Grassland&Field&Wilderness&Forest&Riverside&Gravel&Rural&Asphalt&Cement&Mud &Total &\%\\ 
				\hline
				Train&3,921&	5,610&	2,146&	7,005&	1,770&	4,424&	5,734&	1,467&	7,481&	369&	39,927&69.07\% \\ 
				Val & 345&	960&	343&	478&	334&	642&	860&	336&	1,086&	333&	5,717&9.89\%\\
				Test & 1,092&	1,900&	1,047&	1,785&	393&	1,233&	1,719&	528&	1,999&	468&	12,164&21.04\%\\
				\midrule
				Total&9.27\%&	14.65\%&6.12\%&	16.03\%&	4.32\%&	10.90\%&	14.38\%&	4.03\%&	18.28\%&	2.02\% &57,808&100\% \\
				\hline
			\end{tabular}
			\label{scene}
		}
	\end{center}
\end{table*}

\begin{table}[t]
	\caption{Distribution of weather conditions in ORAD-3D.}
	\begin{center}
		\resizebox{0.41\textwidth}{!}{
			\begin{tabular}{c|cccc|c}
				\hline
				Split&Sun & Rain & Fog & Snow &Total  \\ 
				\hline
				Train& 15,737&	11,741&	8,863&	3,586&39,927\\
				Val & 2,976&	679	&1,450&	612&5,717\\
				Test & 5,209&	2,654&	3,958&	343&12,164\\
				\midrule
				Total &
				41.38\% &	26.08\% &	24.69\% &	7.86\% &57,808\\ 
				\hline
			\end{tabular}
			\label{weather_con}
		}
	\end{center}
\end{table}
\begin{figure}[t]
	\centering
	\centerline{\includegraphics[width=3.4in]{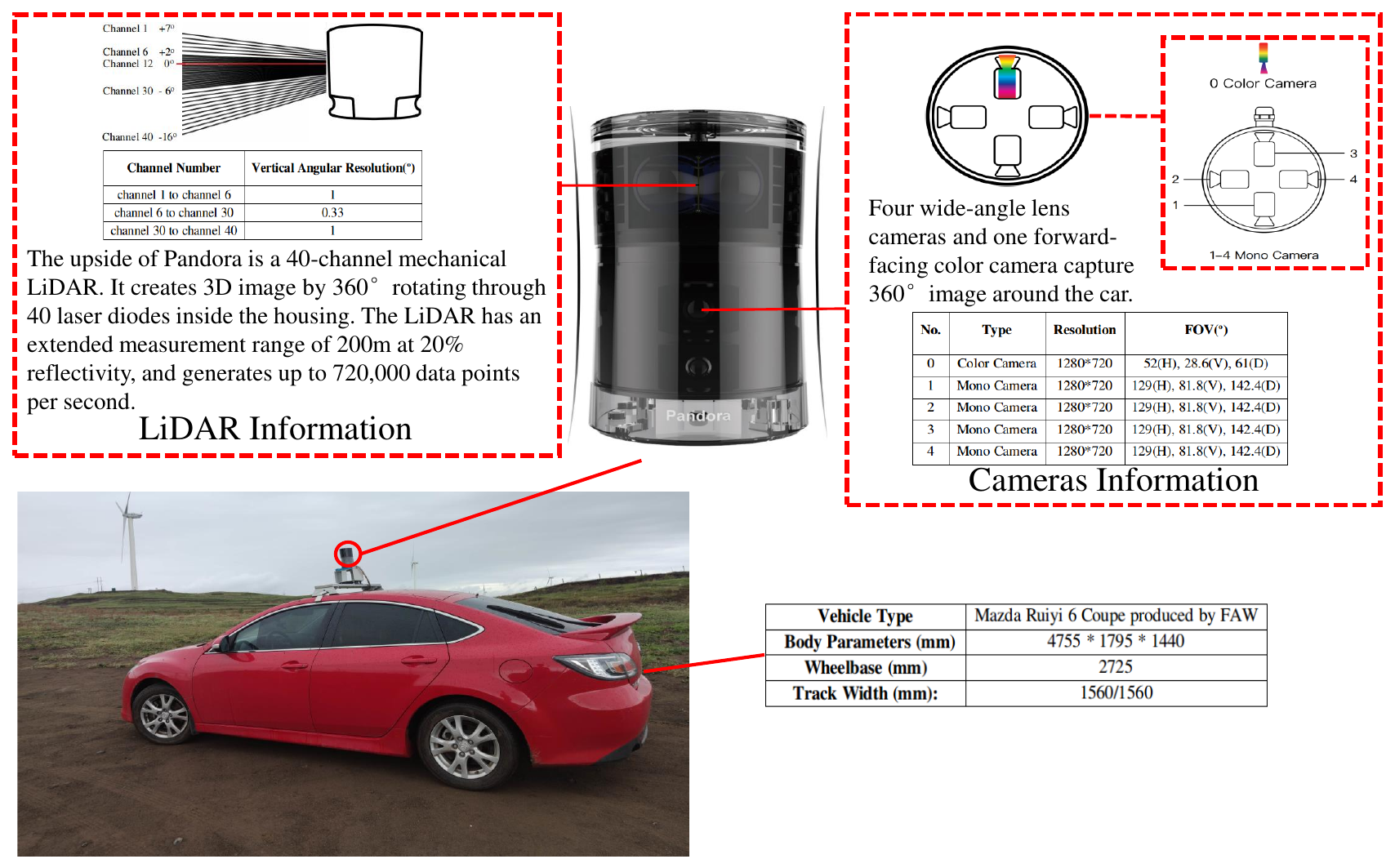}}
	\caption{Detail information of vehicle and sensor to collect LiDAR and camera data.}
	\label{sensor}
\end{figure}
\begin{table}[t]
	\caption{Distribution of lighting variations in ORAD-3D.}
	\begin{center}
		\resizebox{0.45\textwidth}{!}{
			\begin{tabular}{c|cccc|c}
				\hline
				Split&Bright
				light&Daylight & Twilight & Darkness &Total  \\ 
				\hline
				Train &7,515 &15,536&	6,840&	10,036&	39,927\\
				Val &690 &2,758&	1,599&	670&	5,717\\
				Test &2,938 &5,964&	550	&2,712&	12,164\\
				\midrule
				Total&19.28\%&41.96\%&	15.55\%&	23.21\%&57,808\\ 
				\hline
			\end{tabular}
			\label{light}
		}
	\end{center}
\end{table}

\subsection{Data Curation and Statistics}\label{data-description}
\subsubsection{Diverse Scenes}
Unlike the uniformity of structured on-road environments, off-road scenarios are characterized by diverse types, such as grasslands, forests, deserts, farmlands, and mountainous terrain, as shown in Fig.~\ref{type}. Furthermore, off-road environments contain numerous elements that are irregular in shape, with blurred boundaries and semantically ambiguous categories. The heterogeneity and spatial disorder of open off-road scenes significantly exceed the cognitive limits of structured roadways. Structurally, the absence of lane markings and traffic signs removes clear guidance, while natural features like trees and deep ravines contribute to feature sparsity in models. To address these challenges, we have collected off-road data from a variety of environments, including grasslands, forests, deserts, and farmlands, laying the foundation for off-road autonomous driving. The data distribution across various scenes is shown in Table~\ref{scene}.

\subsubsection{Weather Conditions}
Off-road scenarios often encounter sudden weather changes, such as shifts from sunny to rainy, foggy to snowy conditions, and so on, as shown in Fig.~\ref{weather}. Typically, there is a larger volume of autonomous driving data available for sunny weather. However, obtaining corner case data for rare and extreme weather scenarios, such as rain, fog, and snow, is crucial for enhancing the accuracy of off-road autonomous driving models. To address this, we have specifically collected off-road scene data under various extreme weather conditions, including rain, snow, fog, and sunny days. The data distribution across weather conditions is shown in Table~\ref{weather_con}.

\subsubsection{Lighting Variations}
Lighting variations have a significant impact on autonomous driving performance. To address this, we have collected off-road scene data under various lighting conditions, including bright light, daylight, twilight, and darkness, as shown in Fig.~\ref{day}. Data collection during the evening and nighttime is particularly challenging due to reduced visibility and the complexity of environmental factors. To overcome these challenges, we specifically focused on gathering data during twilight and nighttime, allowing the model to learn the unique distribution of off-road data under low-light conditions. By incorporating these varying lighting conditions into the training process, we aim to improve the robustness and accuracy of off-road autonomous driving models in challenging lighting situations. The data distribution across lighting variations is shown in Table~\ref{light}.

\subsubsection{Road Type}
Autonomous vehicles are required to perform a variety of driving maneuvers, such as traveling in a straight line, turning, and navigating complex terrain. Generating video representations of future scenes based on these driving actions is highly valuable, as it allows the vehicle to anticipate and adapt to its surroundings. In off-road environments, the terrain is often rugged and uneven, with frequent uphill and downhill maneuvers that present significant challenges to both vehicle control and data collection. This makes it difficult to obtain comprehensive and diverse datasets that capture the full range of possible off-road driving scenarios. 
To address this challenge, we have collected a diverse set of video data corresponding to various driving actions, with a particular focus on off-road scenarios, as illustrated in Fig.~\ref{turn}. 

\subsubsection{Data collection}
The vehicle used for collecting the ORAD-3D dataset is the Mazda Ruiyi 6 Coupe, manufactured by FAW. Its body dimensions are 4,755 mm $\times$ 1,795 mm $\times$ 1,440 mm, with a wheelbase of 2,725 mm and a track width of 1,560 mm (both front and rear). A sensor fusion kit, Pandora, developed by Hesai Technology, is mounted on the vehicle's roof to capture LiDAR point cloud and RGB image data. The system is equipped with five cameras positioned around the lower part of the vehicle, including one color camera and four wide-angle black-and-white cameras. Detailed specifications of the vehicle and sensors are shown in Fig.~\ref{sensor}. We collected 145 sequences from various off-road environments across China, spanning from spring to winter. Each sequence covers a distance of approximately 100 meters, and the size of the RGB images is 1280 $\times$ 720 pixels.

\section{ORAD-3D Benchmarks}

Building on the large-scale off-road autonomous driving dataset ORAD-3D, we evaluate five core tasks: (1) 2D free-space detection, (2) 3D occupancy prediction, (3) rough GPS-guided path planning, (4) VLM-based autonomous driving, and (5) off-road world model.

\subsection{2D Free-Space Detection}\label{tasks}
\begin{figure}[h]
	\centering
	\centerline{\includegraphics[width=3.4in]{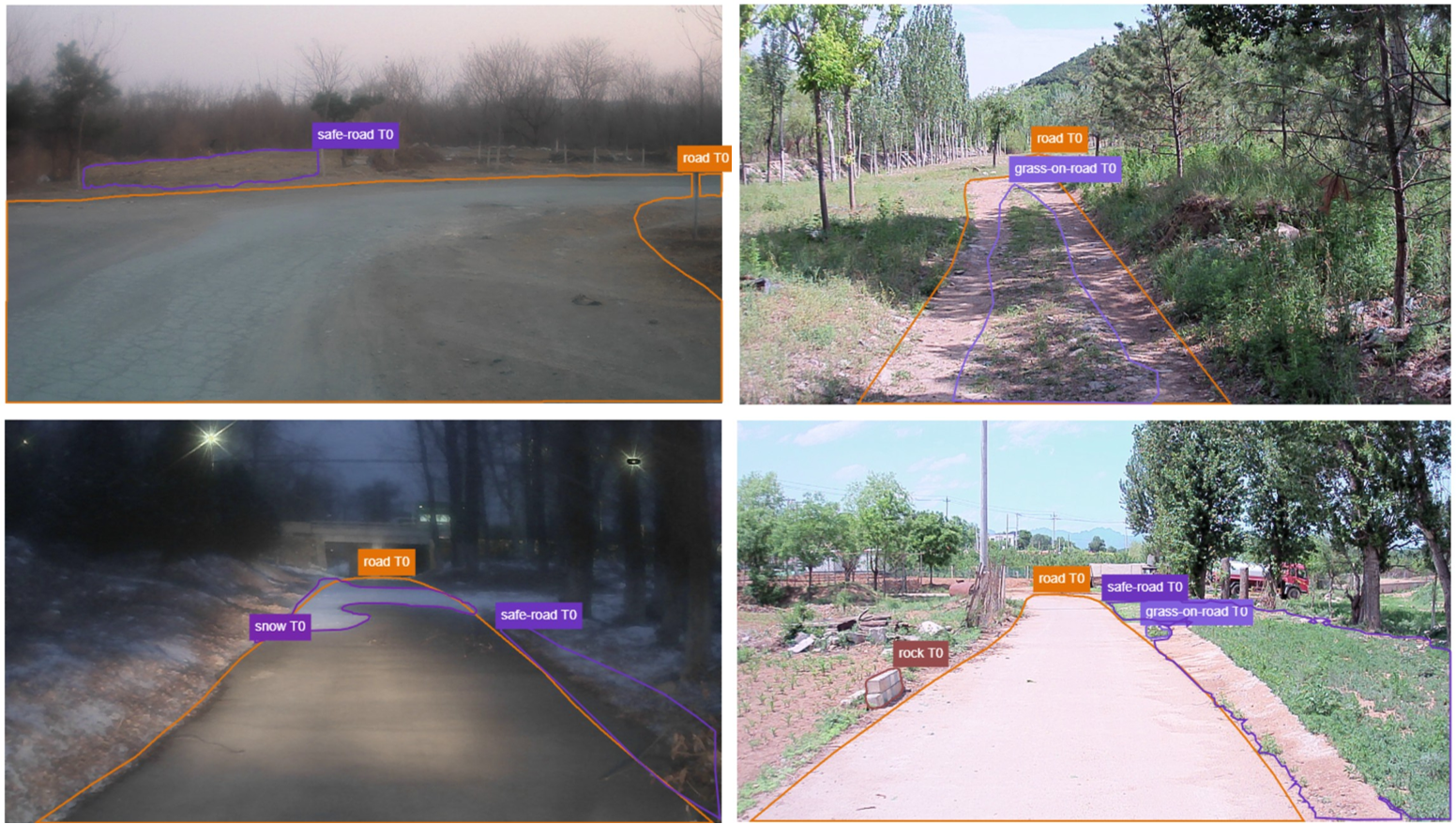}}
	\caption{The annotation 2D off-road free-space detection.}
	\label{orfd}
\end{figure}
\subsubsection{Label Generation}
Unlike the ORFD~\cite{orfd} dataset, which adopts a binary road/non-road labeling scheme, our annotation protocol (Fig.~\ref{orfd}) captures the complexity of off-road environments by including fine-grained classes such as safe road, boundary transition zones, puddles, rocks, vehicles, and pedestrians. This detailed semantic representation enables more nuanced terrain understanding and supports safer autonomous navigation.

\subsubsection{Benchmark Method}
Building on existing off-road free-space detection methods, we conduct experiments on the ORAD-3D dataset using both multimodal and vision-only approaches. Detailed comparative analyses are presented in the experimental results section.

\subsection{3D Occupancy Prediction}
\begin{figure}[h]
	\centering
	\centerline{\includegraphics[width=3.4in]{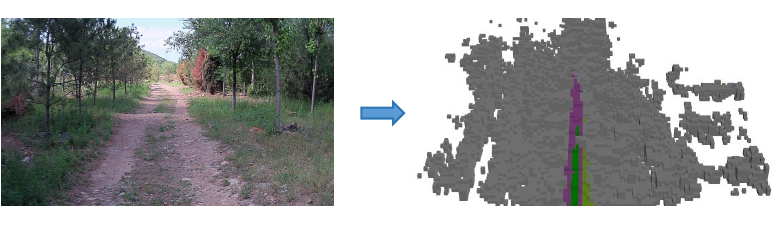}}
	\caption{The Annotation 3D occupancy prediction.}
	\label{occ}
\end{figure}

\subsubsection{Label Generation}
Relying solely on 2D off-road free-space detection results is insufficient for accurately modeling complex 3D terrains. To address this limitation, we further construct 3D occupancy annotations for off-road environments. Specifically, we first apply the LiDAR odometry method KISS-ICP~\cite{vizzo2023ral} to register the collected LiDAR point clouds and obtain precise pose estimates. Multiple frames are then accumulated to generate dense point clouds, from which the 3D occupancy labels are subsequently derived (Fig.~\ref{occ}).

Existing off-road 3D occupancy prediction benchmarks, such as Wild-Occ~\cite{zhai2024wildocc}, are collected from small-scale platforms with limited coverage and relatively simple terrain.

\subsubsection{Benchmark Method}
We conduct extensive experiments on ORAD-3D using a range of 3D occupancy prediction methods, encompassing both general-purpose approaches and algorithms tailored for off-road environments.

\subsection{Rough GPS-guided Path Planning}

\begin{figure}[h]
	\centering
	\centerline{\includegraphics[width=3.4in]{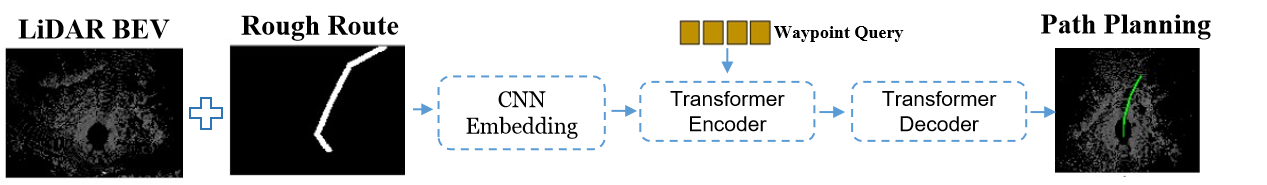}}
	\caption{The flowchart for rough GPS-guided local path planning.}
	\label{model}
\end{figure}

\subsubsection{Label Generation}
In off-road environments, GPS signals are often unreliable, preventing autonomous vehicles from obtaining accurate localization. To address this, we construct a rough GPS-guided path planning benchmark dataset. Specifically, we use the previously estimated poses as ground-truth driving trajectories and apply B-spline interpolation to generate waypoints. We then introduce controlled perturbations to these trajectories to simulate the effects of inaccurate GPS localization.

\subsubsection{Benchmark Method}

We adopt the method proposed by Xu et al.~\cite{xu2022trajectory} as our benchmark for rough GPS-guided path planning (Fig.~\ref{model}). This approach takes LiDAR BEV maps and rough routes as inputs, employing a Transformer-based backbone to predict waypoints. Building upon this, we further incorporate an uncertainty module to enhance the accuracy and reliability of the path planning.

\subsection{VLM-based Autonomous Driving}\label{tass2}
\begin{figure}[h]
	\centering
	\centerline{\includegraphics[width=3.4in]{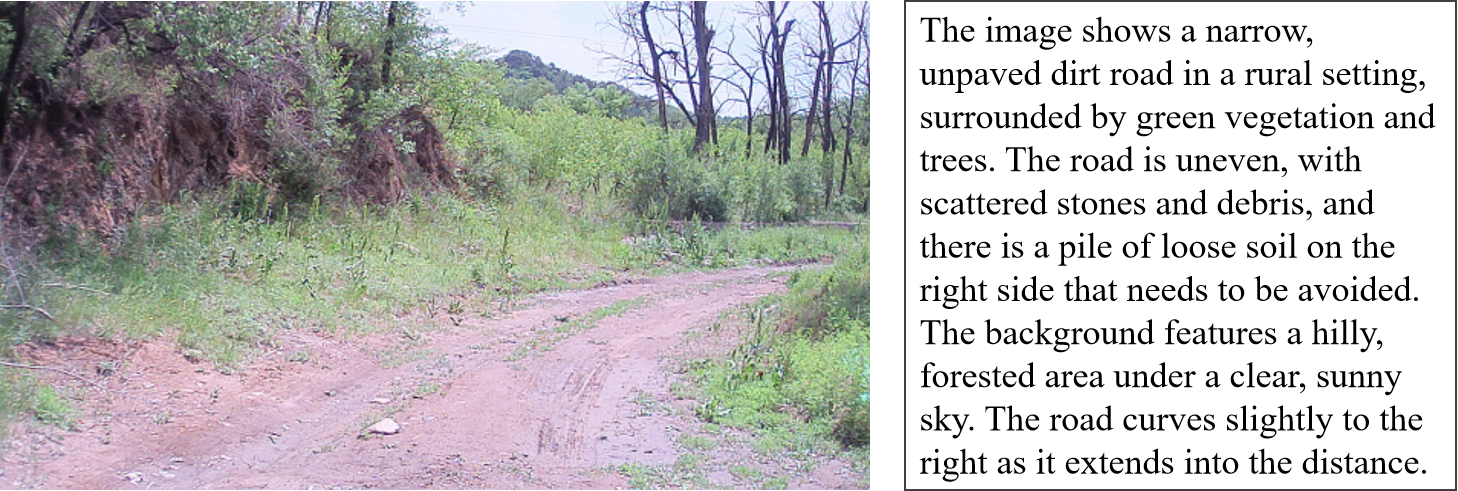}}
	\caption{Textual description annotation of off-road scenes.}
	\label{dis}
\end{figure}
\subsubsection{Label Generation}
Recent advances in VLMs have demonstrated their ability to analyze complex scenes through chain-of-thought reasoning, enabling end-to-end autonomous driving. However, research on applying VLMs to off-road autonomous driving remains limited. In this work, we first leverage the multimodal large model Qwen2.5-VL~\cite{Qwen2.5-VL} to annotate images from the ORAD-3D dataset, generating detailed scene descriptions as illustrated in Fig.~\ref{dis}. Subsequently, using pose data as ground-truth trajectories, we prompt the VLM to predict future paths, enabling end-to-end path planning in off-road scenarios.

\subsubsection{Benchmark Method}
\begin{figure}[h]
	\centering
	\centerline{\includegraphics[width=3.4in]{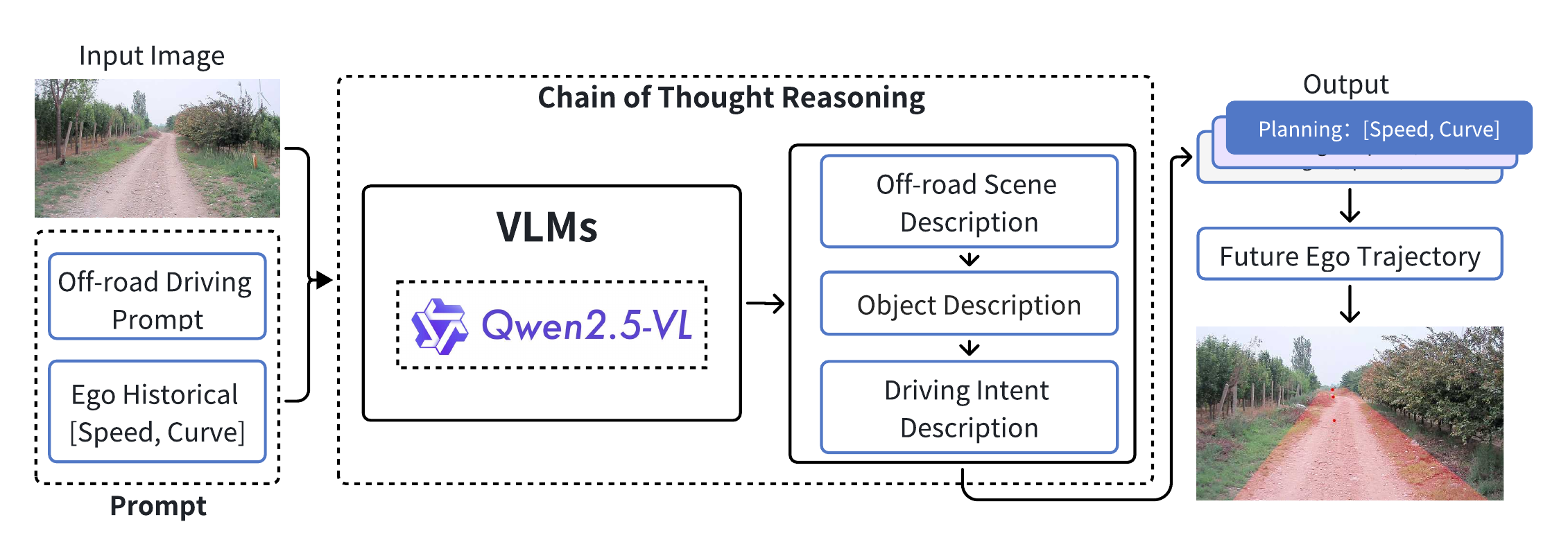}}
	\caption{The flowchart for VLM-based off-road autonomous driving.}
	\label{vlm}
\end{figure}
We construct an off-road VLM-based autonomous driving benchmark on OpenEMMA~\cite{xing2025openemma}, integrating a chain-of-thought (CoT) reasoning mechanism to handle the complexity of unstructured environments. As shown in Fig.~\ref{vlm}, the model takes RGB images and prompt as inputs, performs step-by-step analysis of terrain, obstacles, and navigable areas, and outputs future trajectory waypoints for end-to-end path planning, leveraging VLM generalization while adapting reasoning to off-road challenges.

\subsection{Off-road World Model}
In contrast to the abundance of large-scale datasets readily available for urban autonomous driving, research on off-road autonomous driving remains significantly constrained by the scarcity of suitable data. This paper seeks to address this critical limitation by investigating off-road scene data generation through the development of a world model capable of producing diverse and controllable off-road scenarios~\cite{zhu2024sora}. Specifically, the proposed approach enables the synthesis of off-road data under extreme weather and illumination conditions, from multiple viewpoints, and across heterogeneous road environments, thereby substantially enhancing both the scale and diversity of off-road datasets.

\subsubsection{Benchmark Method}
\begin{figure}[h]
	\centering
	\centerline{\includegraphics[width=3.4in]{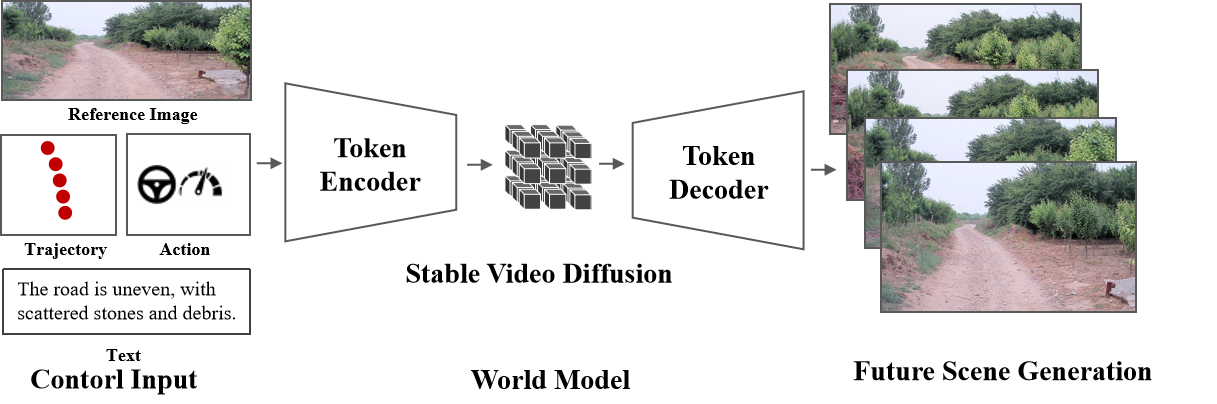}}
	\caption{The flowchart for off-road world model.}
	\label{wm}
\end{figure}
We introduce an off-road world model framework (Fig.~\ref{wm}) that redefines the architecture of Stable Video Diffusion (SVD)~\cite{svd} for off-road autonomous driving applications. Diverging from conventional image-to-video generation paradigms, our approach introduces a dual-stream conditioning mechanism that simultaneously processes visual observations and vehicular control signals~\cite{wang2024drivingdojo}. These features are then spatially aligned with the encoded visual features from the initial observation frame using cross-attention layers in the modified U-Net backbone~\cite{ronneberger2015u}. This synergistic integration enables the diffusion model to generate temporally coherent video predictions that strictly adhere to specified conditions while maintaining visual consistency with the environmental context.

\section{Experimental Results}
In this section, experiments are conducted to validate the performance of the proposed dataset and baseline method.


\subsection{2D Free-Space Detection} 
\begin{figure}[h]
	\centering
	\centerline{\includegraphics[width=3.4in]{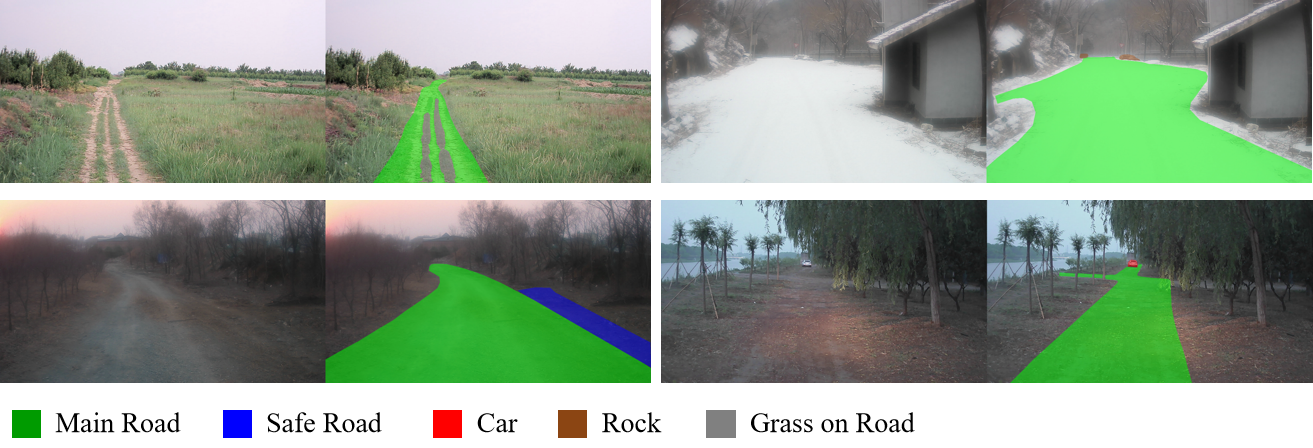}}
	\caption{Visualization of 2D free-space detection.}
	\label{road_re}
\end{figure}

\begin{table}[h]
	\caption{Quantitative results on the ORAD-3D testing set}
	\begin{center}
		\setlength{\tabcolsep}{1.5mm}{
			\begin{tabular}{c|c|c|c}
				\hline
				Method &OFF-Net~\cite{orfd}&M2F2-Net~\cite{m2f2}& ROD~\cite{sun2025rod}  \\ 
				\hline
				mIoU$\uparrow$  &  62.4\% & 65.9\% & 70.3\%  \\
				\bottomrule
			\end{tabular}
			\label{results_road}
		}
	\end{center}
\end{table}

As shown in Fig.~\ref{road_re}, the proposed off-road free-space detection algorithm (ROD~\cite{sun2025rod}) effectively delineates off-road boundaries, median grass strips, and safe regions, thereby providing fine-grained road information that facilitates off-road autonomous driving. Quantitative results are presented in Table~\ref{results_road}. Compared to multimodal baselines such as M2F2-Net~\cite{m2f2} and OFF-Net~\cite{orfd}, the vision-only ROD~\cite{sun2025rod}, built upon a ViT backbone, demonstrates superior accuracy in predicting detailed road information.

\subsection{3D Occupancy Prediction}
\begin{figure}[h]
	\centering
	\centerline{\includegraphics[width=3.4in]{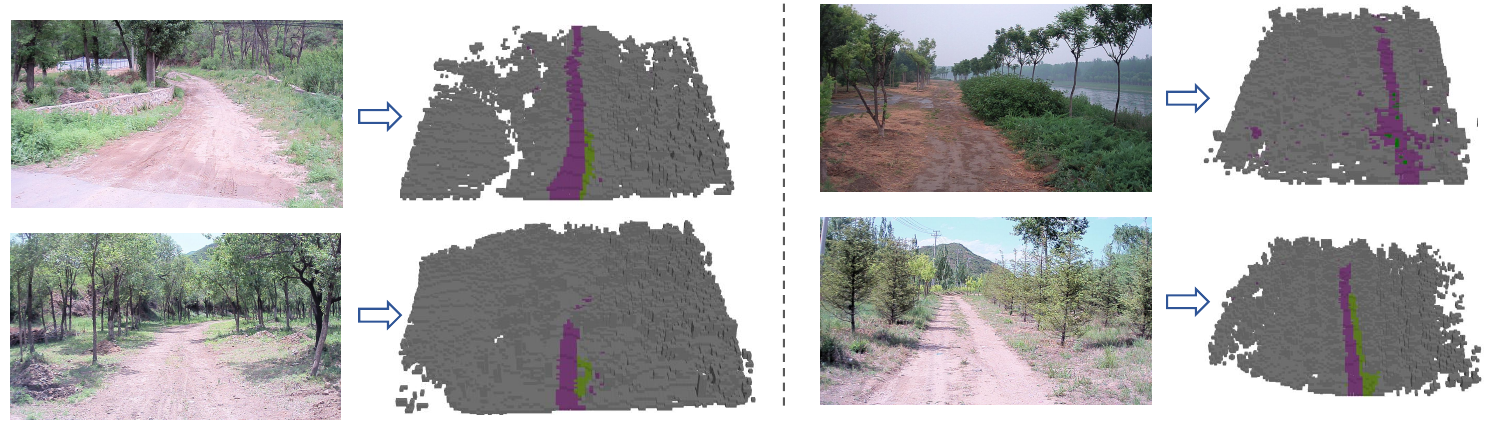}}
	\caption{Visualization of 3D occupancy prediction.}
	\label{occ_re}
\end{figure}

\definecolor{dirt}{rgb}{0.43, 0.08, 0.54}
\definecolor{grass}{rgb}{0, 0.5, 0}
\definecolor{tree}{rgb}{0, 1, 0} 
\definecolor{bush}{rgb}{1, 0.6, 0.8} 
\definecolor{Barrier}{rgb}{0.16, 0.5, 1} 
\definecolor{Puddle}{rgb}{0.52, 1, 0.94} 
\definecolor{Mud}{rgb}{0.39, 0.26, 0.13} 

\begin{table}[h!]
	\caption{3D Semantic Occupancy Prediction Results on ORAD-3D test set.}
	\centering
		\begin{tabular}{c|cc}
			\toprule
			Method& IoU$\uparrow$& mIoU$\uparrow$ \\
			\midrule
			OpenOcc\cite{openocc} & 9.54 & 5.83\\
			OccFormer\cite{Zhang2023OccFormerDT}& 12.24 & 7.45\\
			SurroundOcc~\cite{surroundocc}& 11.87 & 7.60\\
			\bottomrule
	\end{tabular}
	\label{tab:comparisons}
\end{table}

Accurate 3D terrain modeling is crucial for off-road autonomous driving. As shown in Fig.~\ref{occ_re}, the predicted 3D occupancy results demonstrate that the model can capture the 3D structure of complex off-road environments. Quantitative results are reported in Table~\ref{tab:comparisons}, where the fusion of LiDAR and vision yields higher prediction accuracy for 3D occupancy estimation.

\subsection{Rough GPS-guided Path Planning}
\begin{figure}[h]
	\centering
	\centerline{\includegraphics[width=3.4in]{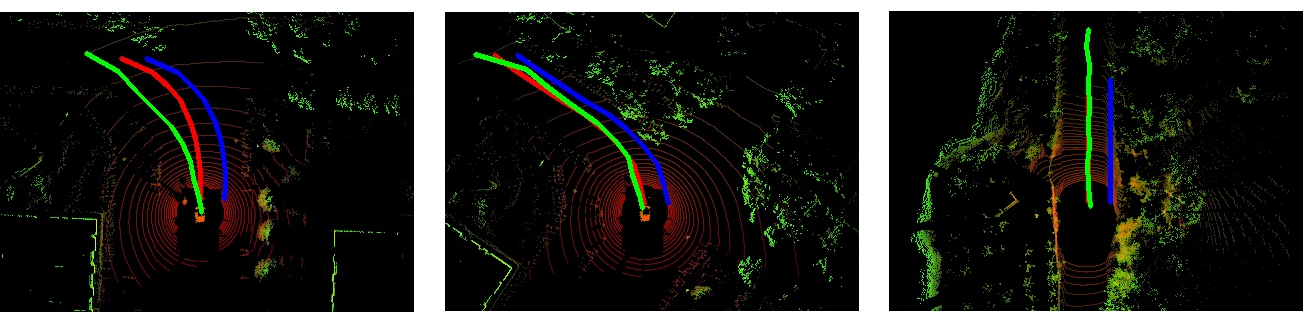}}
	\caption{Visualization of local path planning under the guidance of rough GPS. Blue: local route. Red: ground truth trajectory. Green: predicted trajectory.}
	\label{traj_re}
\end{figure}
\begin{table}[h]
	\caption{Rough GPS-guided Path Planning.}
	\begin{center}
		\setlength{\tabcolsep}{2.5mm}{
			\begin{tabular}{c|ccc}
				\hline
				Method &FDE$\downarrow$& minADE$_{1}$$\downarrow$ & HitRate$_{1,2m}$$\uparrow$   \\ 
				\hline
				Xu et al.~\cite{xu2022trajectory} & 2.34 & 1.07&0.55 \\ 
				\midrule
				Baseline &  2.12 & 0.93 & 0.63\\  
				\bottomrule
			\end{tabular}
			\label{gps_re}
			}
	\end{center}
\end{table}
As shown in Fig.~\ref{traj_re}, although GPS localization is imprecise, the model can effectively plan trajectories along the safe road centerline under the guidance of rough GPS, particularly in turning regions. This capability is crucial for off-road autonomous driving, where localization is often unreliable. Table~\ref{gps_re} provides the quantitative comparison of trajectory planning performance.
\subsection{VLM-based Autonomous Driving}
\begin{figure}[h]
	\centering
	\centerline{\includegraphics[width=3.4in]{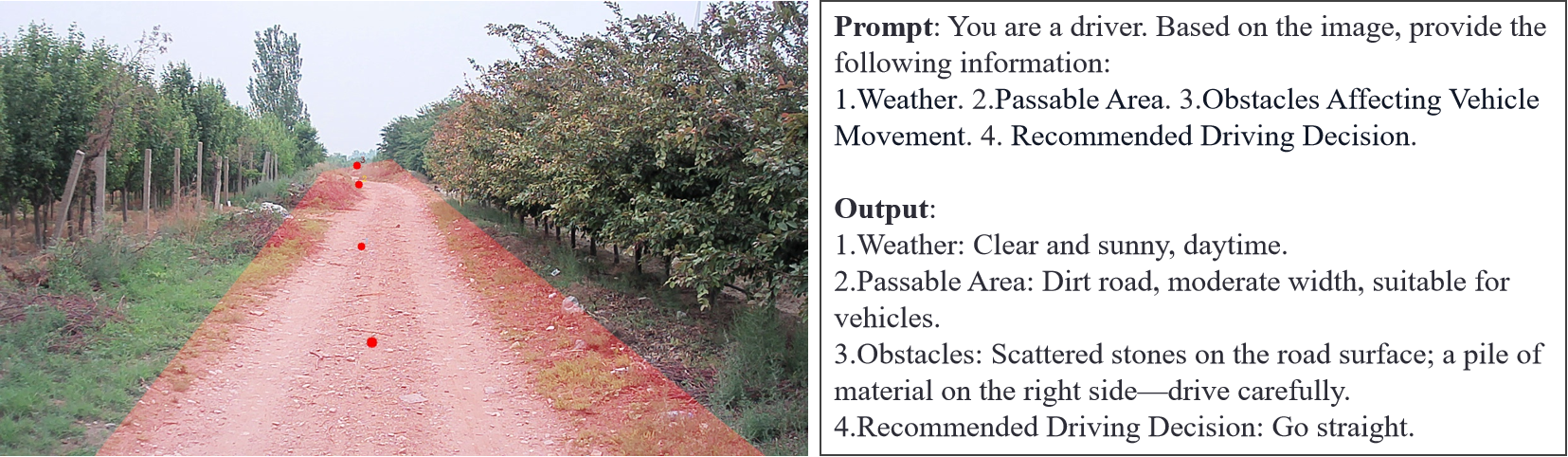}}
	\caption{Visualization of the predicted trajectories via VLM.}
	\label{vla_vlm}
\end{figure}
\begin{table}[h]
	\centering
	\caption{VLM-based Path Planning.}
	\resizebox{0.5\textwidth}{!}
	{
		\begin{tabular}{c|cccc|c}
			\toprule
			\multirow{2}*{Method} &\multicolumn{4}{c|}{L2(m)$\downarrow$} &\multirow{1}*{Failure}\\
			&1s &2s &3s &Avg. &Rate(\%)$\downarrow$\\
			\midrule
			Qwen2.5-VL~\cite{Qwen2.5-VL}  &1.95&2.21&2.90&2.35&82.55\\
			OpenEMMA~\cite{xing2025openemma}  &1.73 &1.92 &2.75 &2.13 &57.56 \\
			LightEMMA~\cite{qiao2025lightemma} &1.72 &1.88 &2.63 &2.08 &55.63 
			\\
			\bottomrule
		\end{tabular}
		\label{vla}
	}
	
	\label{tab:planning}
\end{table}
As illustrated in Fig.~\ref{vla_vlm}, VLMs can perform fine-grained analysis of off-road scenes conditioned on prompts, such as recognizing objects, weather, and road conditions, and subsequently generating a planned trajectory. Quantitative results in Table~\ref{vla} demonstrate that OpenEMMA~\cite{xing2025openemma} and LightEMMA~\cite{qiao2025lightemma} outperform the zero-shot Qwen2.5-VL~\cite{Qwen2.5-VL} in path planning tasks.
\subsection{Off-road World Model}
\begin{figure}[h]
	\centering
	\centerline{\includegraphics[width=3.3in]{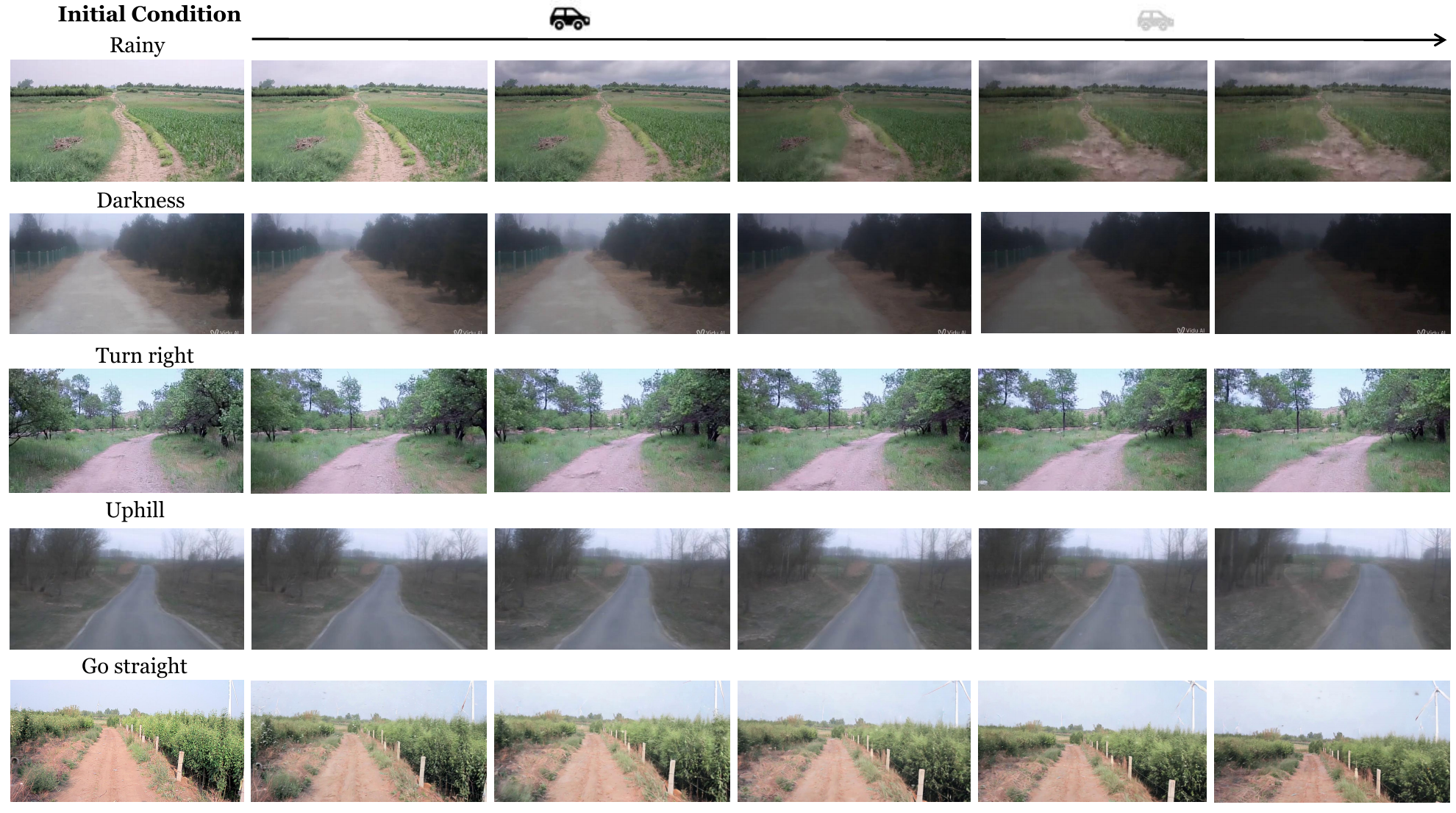}}
	\caption{Future driving video generation will involve the interaction of driving actions with various conditions, enabling the creation of driving videos that correspond to different scenarios and driving behaviors.}
	\label{result1}
\end{figure}
\begin{table}[h]
	\caption{Quantitative results of off-road world model.}
	\begin{center}
			\begin{tabular}{c|c|cc}
				\hline
				Method & Dataset &FID$\downarrow$ & FVD$\downarrow$ \\ 
				\hline
				SVD~\cite{svd}&-&79.5 &534.2  \\
				Baseline&ORAD-3D&70.2 &486.6 \\   
				\hline
			\end{tabular}
			\label{wm_re}
	\end{center}
\end{table}
The results in Table~\ref{wm_re} demonstrate that models trained on our dataset deliver superior visual quality. As shown in Fig.~\ref{result1}, the world model trained on the ORAD-3D dataset is capable of accurately generating future scene videos. Notably, the model can generate the required future scenarios based on specific conditions, greatly reducing the cost of collecting off-road scene data and advancing research in off-road autonomous driving.

\section{Conclusion}

In this work, we introduced ORAD-3D, the largest and most diverse dataset for off-road autonomous driving to date. Collected across a wide variety of terrains, weather conditions, and illumination scenarios, ORAD-3D provides multi-sensor data to support both 2D and 3D perception tasks. Beyond data collection, we established four representative benchmarks—2D free-space detection, 3D occupancy prediction, rough GPS-guided path planning, VLM-based autonomous driving, and off-road world model—covering essential challenges in perception and decision-making for unstructured environments. We believe ORAD-3D will serve as a key resource to advance research on robust off-road autonomy, facilitating the development of algorithms capable of operating reliably under diverse and adverse conditions.

\textbf{Limitations and Future Work.}
Off-road environments are highly complex, yet current datasets remain limited in scale and obstacle diversity. Future work will expand terrain coverage and capture more challenging scenarios with diverse obstacles to better evaluate algorithm robustness in real-world conditions.
\bibliographystyle{IEEEtran}
\bibliography{references}

\end{document}

%% file: shortcuts.tex
\usepackage{color}
\definecolor{red}{rgb}{1.00,0.20,0.20}
\definecolor{blue}{rgb}{0.20,0.20,1.00}
\definecolor{green}{rgb}{0.00,1.00,0.00}

\usepackage{array}
\newcolumntype{L}[1]{>{\raggedright\let\newline\\\arraybackslash\hspace{0pt}}m{#1}}
\newcolumntype{C}[1]{>{\centering\let\newline\\\arraybackslash\hspace{0pt}}m{#1}}
\newcolumntype{R}[1]{>{\raggedleft\let\newline\\\arraybackslash\hspace{0pt}}m{#1}}


%% file: acronyms.tex
\ifdefined\UseShortAcronyms

\else

\fi